\documentclass[10pt,twocolumn,letterpaper]{article}

\usepackage{icb}
\usepackage{times}
\usepackage{epsfig}
\usepackage{graphicx}
\usepackage{amsmath}
\usepackage{amssymb}

\usepackage{url}
\usepackage{hyperref}
\usepackage{breakurl}
\usepackage{multirow}
\usepackage{array}
\usepackage{tabu}
\usepackage{xcolor}

\usepackage[hang,flushmargin]{footmisc}

\icbfinalcopy 

\ificbfinal\fi
\begin{document}

\title{Fingerprint Presentation Attack Detection: Generalization and Efficiency}

\author{Tarang Chugh and Anil K. Jain\\
Department of Computer Science and Engineering\\
Michigan State University, East Lansing, Michigan 48824\\
{\tt\small \{chughtar, jain\}@cse.msu.edu}}

\maketitle
\thispagestyle{empty}

\begin{abstract}
We study the problem of fingerprint presentation attack detection (PAD) under unknown PA materials not seen during PAD training. A dataset of $5,743$ bonafide and $4,912$ PA images of $12$ different materials is used to evaluate a state-of-the-art PAD, namely Fingerprint Spoof Buster. We utilize 3D t-SNE visualization and clustering of material characteristics to identify a \textit{representative set} of PA materials that cover most of PA feature space. We observe that a set of six PA materials, namely Silicone, 2D Paper, Play Doh, Gelatin, Latex Body Paint and Monster Liquid Latex provide a good representative set that should be included in training to achieve generalization of PAD. 
We also propose an optimized Android app of Fingerprint Spoof Buster that can run on a commodity smartphone (Xiaomi Redmi Note 4) without a significant drop in PAD performance (from TDR = $95.7\%$ to $95.3\%$ @ FDR = $0.2\%$) which can make a PA prediction in less than $300$ms.
\end{abstract}
\vspace{-2.9mm}

\section{Introduction}

With the ubiquitous deployment of fingerprint recognition for unlocking smartphones, authenticating financial transactions, international border security, etc., they are now a prime target for presentation attacks. The ISO standard IEC 30107-1:2016(E)~\cite{isopad} defines presentation attacks as the \textit{``Presentation to the biometric data capture subsystem with the goal of interfering with the operation of the biometric system"}. One of the most common ways to realize presentation attacks is using fingerprint spoofs\footnote{Fingerprint spoofs are one of most commonly deployed presentation attacks. Other forms include fingerprint alterations and cadaver fingers.}, \textit{i.e.} \textit{gummy} fingers. \textit{Gummy fingers}~\cite{matsumoto2002impact} refer to counterfeit finger-like objects, fabricated using commonly available materials such as latex, gelatin, wood glue, etc., with an accurate imitation of another individual's friction ridge patterns engraved on its surface. These can be prepared with a multitude of fabrication processes ranging from basic molding and casting\footnote{https://www.youtube.com/watch?v=bp-MrrAmprA} to utilizing sophisticated 2D and 3D printing techniques~\cite{matsumoto2002impact, ghiani2017review, cao2016hacking, engelsma2018universal}. 

\begin{figure}[t]
\centering
\includegraphics[width=0.94\linewidth]{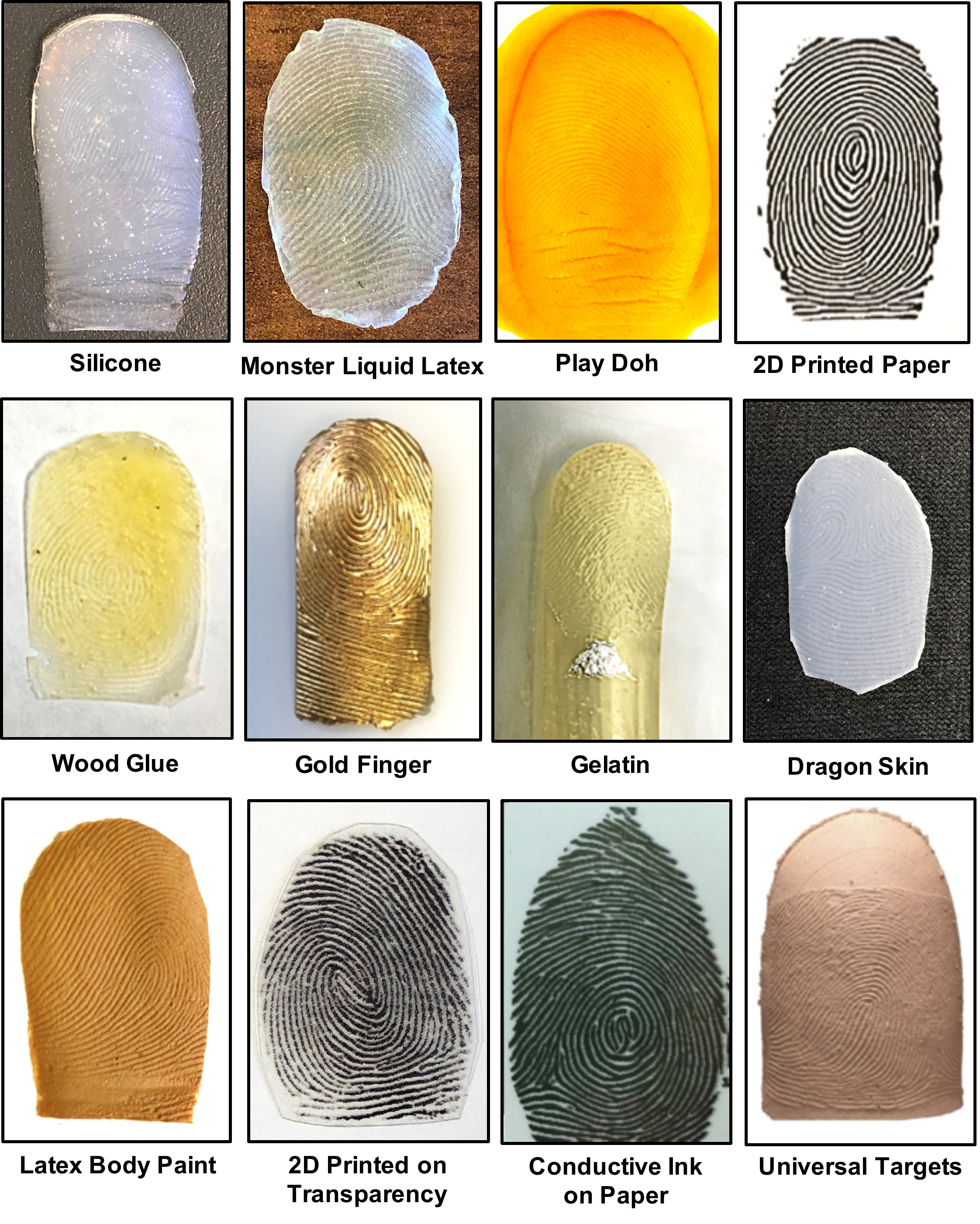}
\caption{Twelve different PA fabrication materials investigated in this study.}
\label{fig:spoofs}
\vspace{-4mm}
\end{figure}

\begin{figure*}[htbp]
\centering
\includegraphics[width=0.91\linewidth]{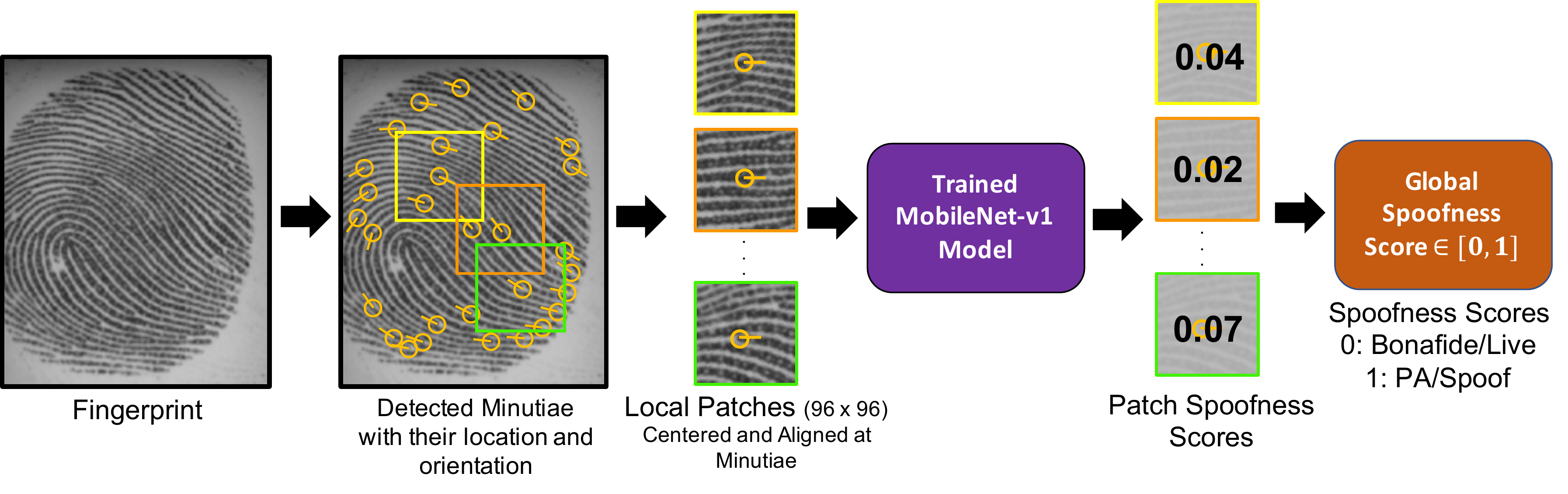}
\caption{An overview of the Fingerprint Spoof Buster~\cite{chugh2018fingerprint}, a state-of-the-art fingerprint PAD, utilizing CNNs trained on local patches centered and aligned using minutiae location and orientation, respectively.}
\label{fig:spoofbuster}
\vspace{-2.5mm}
\end{figure*}


Fingerprint spoof attacks 
were reported to bypass the fingerprint system security with a success rate of more than 70\%~\cite{biggio2012security}. For instance, in March 2013, a Brazilian doctor was arrested for fooling the biometric attendance system for colleagues at a hospital in Sao Paulo using silicone spoof fingers\footnote{http://www.bbc.com/news/world-latin-america-21756709}. In another incident, in Sept. 2013, Germany's Chaos Computer Club hacked the capacitive sensor of recently released Apple iPhone 5s with inbuilt TouchID fingerprint technology by utilizing a high resolution photograph of the enrolled user's fingerprint to fabricate a spoof fingerprint with wood glue\footnote{http://www.ccc.de/en/updates/2013/ccc-breaks-apple-touchid}. And more recently, in July 2016, Michigan State University researchers assisted police in a homicide case by unlocking a fingerprint secure smartphone using a 2D printed fingerprint spoof\footnote{https://www.forbes.com/sites/thomasbrewster/2016/07/28/fingerprint-clone-hack-unlocks-murder-victim-samsung-s6-hacks-apple-iphone-galaxy-s7/}. Many such attacks go unreported.

Given the growing possibilities for fingerprint PA attacks, there is now an urgent requirement for presentation attack detection as a first line of defense to ensure the security of a fingerprint recognition system. In response to this, a series of fingerprint Liveness Detection (LivDet) competitions~\cite{ghiani2017review} have been held since 2009 to advance state-of-the-art and benchmark the proposed PAD solutions, with the latest edition held in 2017~\cite{mura2018livdet}. Generally, presentation attacks can be detected by either: (i) hardware-based approaches, i.e. augmenting the fingerprint readers with sensor(s) to gather evidence of the liveness of the subject, or (ii) software-based approaches, \textit{i.e.} extract features from the presented fingerprint image (or a sequence of frames)~\cite{mura2018livdet, ghiani2017review, marasco2015survey, marcel2014handbook}. In the case of hardware-based approaches, special types of sensors are proposed to detect the characteristics of vitality, such as blood flow~\cite{lapsley1998anti}, skin distortion~\cite{antonelli2006fake}, odor~\cite{baldisserra2006fake}, sub-dermal ridge patterns using multispectral scanner~\cite{tolosana2018towards, rowe2006multispectral} or optical coherence tomography (OCT)~\cite{hogan2015multiple}. An open-source fingerprint reader, that can be constructed using commodity hardware, utilizes a two-camera design providing two complementary streams of information for spoof detection~\cite{engelsma2018fingerprint, engelsma2018raspireader}.

Software-based solutions, unlike hardware-based solutions, work with any commodity fingerprint readers. The various approaches in the literature have explored hand-crafted features: (i) anatomical features (e.g. pore locations and their distribution~\cite{marcialis2010analysis}), (ii) physiological features (e.g. perspiration~\cite{marasco2012combining}), or (iii) texture-based features (e.g. Binarized Statistical Image Features (BSIF)~\cite{ghiani2013fingerprint} and Weber Local Descriptor~\cite{gragnaniello2013fingerprint}). Gragniello et al.~\cite{gragnaniello2015local} utilized spatial and frequency domain information to construct a 2D local contrast-phase descriptor (LCPD) for PAD. 

One of the limitations of the above mentioned PAD approaches is their poor generalization performance against PA materials not seen during training~\cite{rattani2015open}. 
Some studies have modeled PAD as an open-set problem. Rattani et al.~\cite{rattani2015open} utilized Weibull-calibrated SVM (W-SVM), a SVM variant based on properties of statistical extreme value theory, to detect PAs made of novel materials. Ding and Ross~\cite{ding2016ensemble} utilized textural features extracted from only bonafide fingerprint images to train an ensemble of multiple one-class SVMs. 


New approaches to fingerprint PAD have proposed convolutional neural network (CNN) based solutions which have been shown to outperform hand-crafted features on publicly available LivDet databases~\cite{menotti2015deep, nogueira2016fingerprint, jang2017fingerprint, chugh2017fingerprint, pala2017deep, tolosana2018towards, chugh2018fingerprint}. But, there are two major limitations of the CNN based approaches. (i) \textit{Generalization}: the selection of PA materials used in training (known PAs) directly impacts the performance against unknown PAs. It is reported that some PA materials are easier to detect (e.g. EcoFlex, Gelatin, Latex) compared to others (e.g. Wood Glue, Silgum) when left out from training~\cite{chugh2018fingerprint}. (ii) High memory and computation requirements: this inhibits their use in low resource environments such as smartphones or embedded devices (e.g. standalone smart fingerprint readers). Introduction of in-display optical fingerprint technology\footnote{http://www.vivo.com/en/products/v11}$^,$\footnote{https://www.oneplus.com/6t} would require efficient and robust PAD solutions running on the devices.

\begin{table}[t]
\centering
\caption{Summary of the dataset and generalization performance (TDR (\%) @ FDR = $0.2\%$) with leave-one-out method. A total of twelve models are trained where the material left-out from training is taken as the new material for evaluating the model.}
\label{tab:databaseresults}
\resizebox{\linewidth}{!}{
\begin{tabular}{ | >{\centering\arraybackslash}p{2.8cm} | >{\raggedleft\arraybackslash}p{1.3cm} | >{\raggedleft\arraybackslash}p{1.5cm} || >{\centering\arraybackslash}p{2.8cm} | } \hline
\textbf{Fingerprint Presentation Attack Material} & \textbf{\#Images} & \textbf{\#Local Patches} & \textbf{Generalization Performance} \footnotesize{(TDR (\%) @ FDR = 0.2\%)} \\ \hline \hline

\textbf{Silicone} & $1,160$ & $38,145$ & $67.62$ \\ \hline
\textbf{Monster Liquid Latex} & $882$ & $27,458$ & $94.77$ \\ \hline
\textbf{Play Doh} & $715$ & $17,602$ & $58.42$ \\ \hline
\textbf{2D Printed Paper} & $481$ & $7,381$ & $55.44$ \\ \hline
\textbf{Wood Glue} &$397$ & $12,681$ & $86.38$ \\ \hline
\textbf{Gold Fingers} & $295$ & $9,402$ & $88.22$ \\ \hline
\textbf{Gelatin} & $294$ & $10,508$ & $54.95$ \\ \hline 
\textbf{Dragon Skin} & $285$ & $7,700$ & $97.48$ \\ \hline
\textbf{Latex Body Paint} & $176$ & $6,366$ & $76.35$ \\ \hline
\textbf{Transparency} & $137$ & $3,846$ & $95.83$ \\ \hline
\textbf{Conductive Ink on Paper} & $50$ & $2,205$ & $90.00$ \\ \hline
\textbf{3D Universal Targets} & $40$ & $1,085$ & $95.00$ \\ \hline \hline

\textbf{Total PAs} & $\textbf{4,912}$ & $\textbf{144,379}$ & \textbf{\raggedleft{Weighted}} \\ \cline{1-3}

\textbf{Total Bonafide} & $\textbf{5,743}$ & $\textbf{228,143}$ & \textbf{Average:}  $75.24$ \\ \hline
\end{tabular}
}
\vspace{-2mm}
\end{table}

To overcome the above two limitations, Chugh et al.~\cite{chugh2018fingerprint} proposed~\textit{Fingerprint Spoof Buster} utilizing local patches ($96 \times 96$) centered and aligned using fingerprint minutiae to train a MobileNet-v1 model~\cite{howard2017mobilenets}. This fusion of fingerprint domain knowledge (minutiae) and local patches around minutiae provides accuracy and generalization ability of Spoof Buster. We evaluate the performance of Fingerprint Spoof Buster against unknown PAs by adopting a leave-one-out protocol; one material is left out from training set and is then utilized for evaluating cross-material or generalization performance. We utilize 3D t-SNE visualizations of the bonafide and PA samples in the CNN feature space to investigate the relationship between different PA materials and correlate it with their cross-material performances and material characteristics to identify a \textit{representative set} of materials that should be included during training to enhance the generalization performance. We also optimize Fingerprint Spoof Buster for implementation and real-time inference in smartphones. The main contributions of this paper are:

\begin{enumerate}
\item Evaluated the generalization performance of Fingerprint Spoof Buster, a state-of-the-art CNN-based PAD approach using 12 different PA materials.
\item Used 3D t-SNE visualization and material characteristics to identify a ``representative set" of materials (Silicone, 2D paper, Play Doh, Gelatin, Latex Body Paint, and Monster Liquid Latex) that could almost cover the entire PA feature space.
\item Optimized \textit{Fingerprint Spoof Buster} by K-means clustering of minutiae to reduce the number of local patches. Approximately 4-fold savings in computation is achieved while maintaining the PAD performance. 
\item The optimized Android app, called \textit{Fingerprint Spoof Buster Lite}, is implemented using a quantized MobileNet-v1 model. The app accepts a live-scan fingerprint and makes a bonafide v. PA decision in 300 ms on a Xiaomi Redmi Note 4.
\end{enumerate}

\section{Proposed Approach}

\subsection{Fingerprint Spoof Buster}

\textit{Fingerprint Spoof Buster}~\cite{chugh2018fingerprint}, a state-of-the-art PAD approach, utilizes local patches ($96 \times 96$) centered and aligned around fingerprint minutiae. It is trained on a MobileNet-v1~\cite{howard2017mobilenets} CNN model to evaluate its generalization performance. Figure~\ref{fig:spoofbuster} presents an overview of the Fingerprint Spoof Buster. It achieved the state-of-the-art performance~\cite{chugh2018fingerprint} on publicly available LivDet databases~\cite{ghiani2017review} and exceeded the IARPA Odin Project~\cite{IARPAProject} requirement of TDR = $97.0\%$ @ FDR = $0.2\%$.

\subsection{Presentation Attack Database}
\label{sec:database}

We constructed a database of $5,743$ bonafide\footnote{In the literature, the term \textit{live} fingerprint has been primarily used to refer a \textit{bonafide} fingerprint juxtaposed to spoof fingerprints. However, in the context of all forms of presentation attacks, bonafide fingerprint is a more appropriate term as some PAs such as fingerprint alterations also exhibit characteristics of liveness.} and $4,912$ PA images captured on CrossMatch Guardian 200\footnote{https://www.crossmatch.com/wp-content/uploads/2017/05/20160726-DS-En-Guardian-200.pdf}, one of the most popular slap readers. This dataset is created by combining the publicly available MSU Fingerprint Presentation Attack Dataset (MSU-FPAD) and Precise Biometrics Spoof-Kit Dataset (PBSKD)~\cite{chugh2018fingerprint}. Table~\ref{tab:databaseresults} lists the 12 PA materials,  the total number of impressions, and minutiae-based local patches for each material type. Figure~\ref{fig:spoofs} shows fingerprint spoof images fabricated using them.


\begin{figure*}[t]
\centering
\includegraphics[trim=0cm 0cm 0cm 0cm, width=0.91\linewidth]{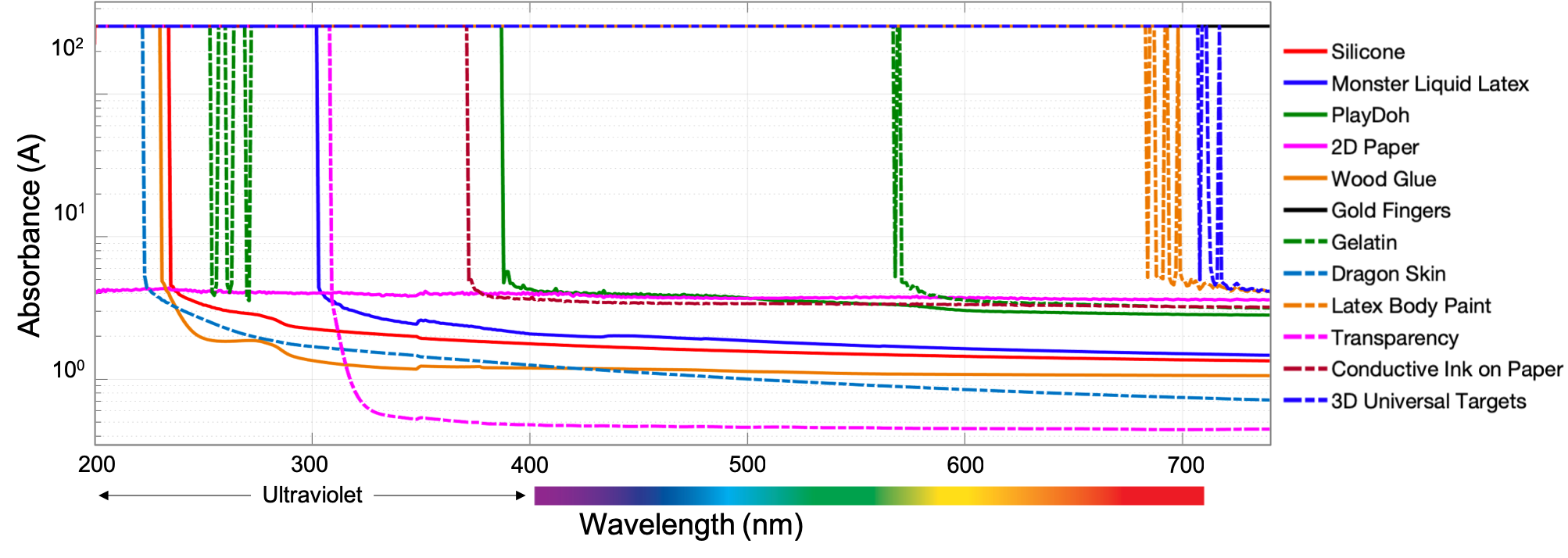}
\caption{Light absorbance property of twelve PA materials in 200nm - 800nm wavelength spectrum.}
\label{fig:uvvis}
\end{figure*}

\begin{figure*}[htbp!]
\centering
\includegraphics[trim=0cm 0cm 0cm 0cm, width=0.91\linewidth]{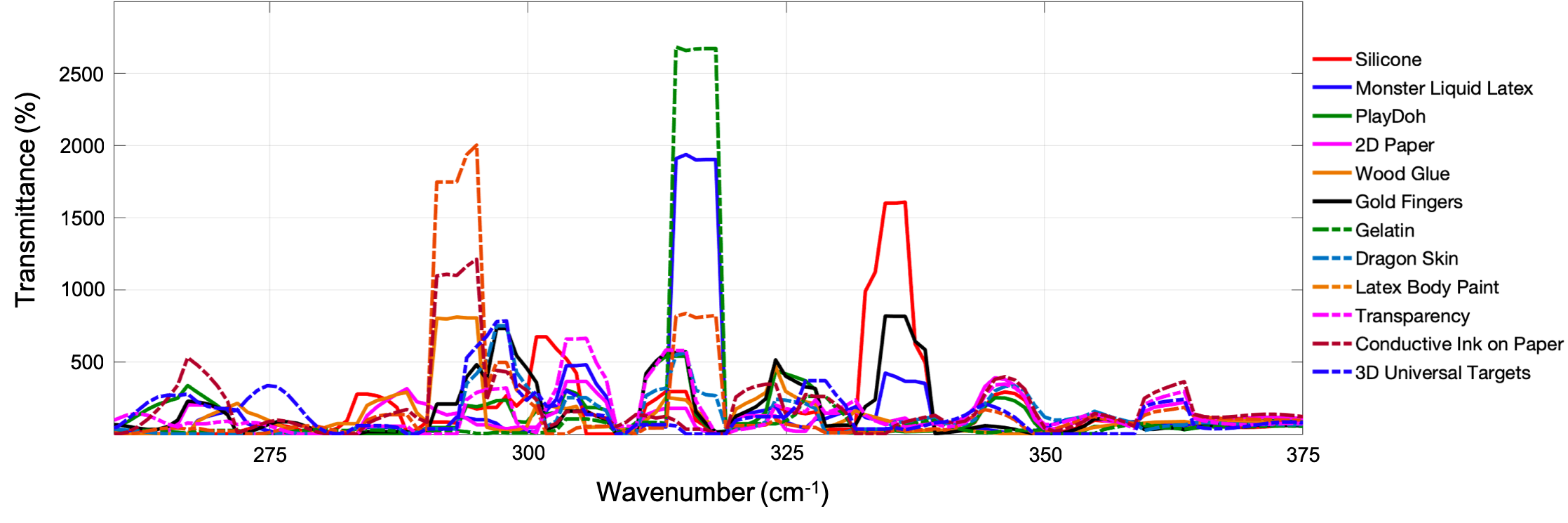}
\caption{Fourier Transform Infrared Spectroscopy of twelve PA materials in the 260 - 375 wavenumber range.}
\label{fig:ftir}
\vspace{-2.6mm}
\end{figure*}

\subsection{Experimental Protocol}

We adopt the leave-one-out protocol to simulate the scenario of encountering \textit{unknown} materials to evaluate the generalization ability of Fingerprint Spoof Buster. One PA material out of the 12 types is left out from the training set which is then utilized during testing. This requires training a total of $12$ different MobileNet-v1 models each time leaving out one of the $12$ different PA types. The $5,743$ bonafide images are partitioned into training and testing such that there are $1,000$ randomly selected bonafide images in testing set and the remaining $4,743$ images in training.

\subsection{Performance against Unknown Materials}

Table~\ref{tab:databaseresults} presents the performance of Fingerprint Spoof Buster against unknown presentation attacks in terms of TDR @ FDR = $0.2\%$. The weighted average generalization performance achieved by the PAD with leave-one-out method is TDR = $75.24\%$, compared to TDR = $97.20\%$ @ FDR = $0.2\%$ when all PA material types are known during training. The PA materials Dragon Skin, Monster Liquid Latex, Transparency, 3D Universal Targets, and Conductive Ink on Paper are easily detected with a TDR $\ge 90\%$ @ FDR = $0.2\%$ even when these materials are not seen by the models during training. On the other hand, PA materials such as PlayDoh, Gelatin, 2D Printed Paper, and Silicone are the most affected by leave-one-out method achieving a TDR $\le 70\%$ @ FDR = $0.2\%$. To understand the reasons for this difference in performance for different materials, we study the material characteristics in the next section.

\subsection{PA Material Characteristics}

Table~\ref{tab:databaseresults} shows that some of the PA materials are easier to detect than others, even when left out from training. To understand the reason for this, it is crucial to identify the relationship between different PAs in terms of their material characteristics. If we can group the PA materials based on shared characteristics, it can be utilized to identify a set of representative materials to train a robust and generalized model. For the given dataset of fingerprint images captured using CrossMatch Guardian 200 optical reader, we measured the following material characteristics: (i) Optical properties: Ultra Violet - Visible (UV-Vis) spectroscopy response and Fourier Transform Infrared (FT/IR) Spectroscopy response, and (ii) Mechanical Properties: material elasticity and moisture content. These material characteristics were selected based on our discussions with material science experts\footnote{Material resistivity would be an important characteristic when performing a similar analysis for capacitive fingerprint readers.}.

\textit{Ultra Violet - Visible (UV-Vis) spectroscopy}: The UV-Vis response of a given material represents the absorption of monochromatic radiations at different wavelengths (ultraviolet (200-400 nm) to visible spectrum (400-750 nm)) by the material. A peak in the UV-Vis response indicates that the material has high absorbance of the light at that given wavelength~\cite{perkampus2013uv}. A Perkin Elmar Lambda 900 UV/Vis/NIR spectrometer\footnote{http://www.perkinelmer.com/category/uv-vis-spectroscopy-uv} was used to measure the light absorbance property of materials shown in Figure~\ref{fig:uvvis}.


\textit{Fourier Transform Infrared (FT/IR) Spectroscopy}: The FT/IR response of a given material is a signature of its molecular structure. The molecules absorb frequencies that are characteristic of their structure, called resonant frequencies, i.e. the frequency of the absorbed radiation matches with the vibrational frequency~\cite{smith2011fundamentals}. An FT/IR signature is a graph of infrared light absorbance (or transmittance) on the Y-axis vs. frequency on the X-axis (measured in reciprocal centimeters i.e. $cm^{-1}$ or wave numbers). Figure~\ref{fig:ftir} presents the FT/IR response of 12 different PA materials measured by Jasco FT/IR-4600 spectrometer\footnote{https://jascoinc.com/products/spectroscopy/ftir-spectrometers/models/ftir-4000-series/}. The FT/IR spectrometer provided material response in the range $250-6,000$ wave numbers, but all the materials exhibited non-zero transmittance only in the range $250-375$ wave numbers. 

\textit{Material Elasticity}: 
A fingerprint spoof fabricated using an elastic material undergoes higher deformation, resulting in large friction ridge distortion when the spoof is pressed against the fingerprint reader's glass platen, compared to less elastic materials. We classify the 12 different PA materials into three classes based on their observed elasticity: (i)~\textit{High elasticity}: Silicone, Monster Liquid Latex, Dragon Skin, Wood Glue, Gelatin, (ii)~\textit{Medium elasticity}:  Play Doh, Latex Body Paint, 3D Universal Targets, and (iii)~\textit{Low elasticity}: 2D Paper, Gold Fingers, Transparency, and Conductive Ink on Paper.

\textit{Moisture Content}: Another crucial material property is the amount of moisture content in it which leads to varying degrees of contrast in the corresponding fingerprint image. PA materials with high moisture content (e.g. Silicone) produce high contrast images compared to materials with low moisture content (e.g. 2D Paper) on CrossMatch reader. We classify the 12 different PA materials into three classes of moisture content level based on the observed image contrast: (i)~\textit{High Moisture Level}: Silicone, Play Doh, Dragon Skin, (ii)~\textit{Medium Moisture Level}: Monster Liquid latex, Wood Glue, Gold Fingers, Gelatin, 3D Universal Targets,  and (iii)~\textit{Low Moisture Level}: 2D Paper, Latex Body Paint, Transparency, Conductive Ink on Paper.

\begin{figure}[t!]
\centering
\includegraphics[trim=0.5cm 0cm 0cm 0cm, width=\linewidth]{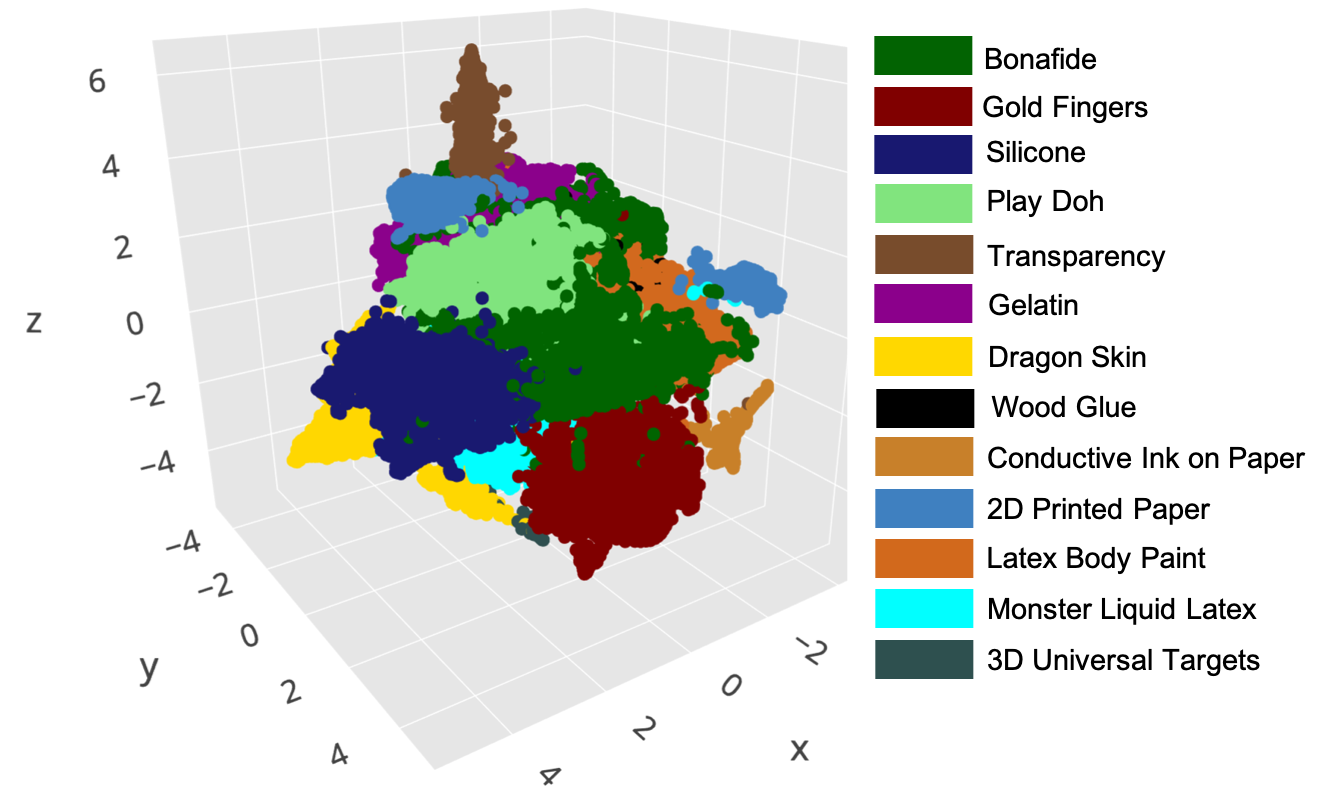}
\caption{Representation of bonafide fingerprints and presentation attack instruments fabricated with different materials in the 3D t-SNE feature space. The original representation is 1024-dimensional obtained form the trained CNN model.}
\label{fig:sampletsne}
\vspace{-4mm}
\end{figure}

\begin{figure}[t]
\centering
\includegraphics[trim=1.6cm 1.1cm 0.5cm 0.5cm, width=\linewidth]{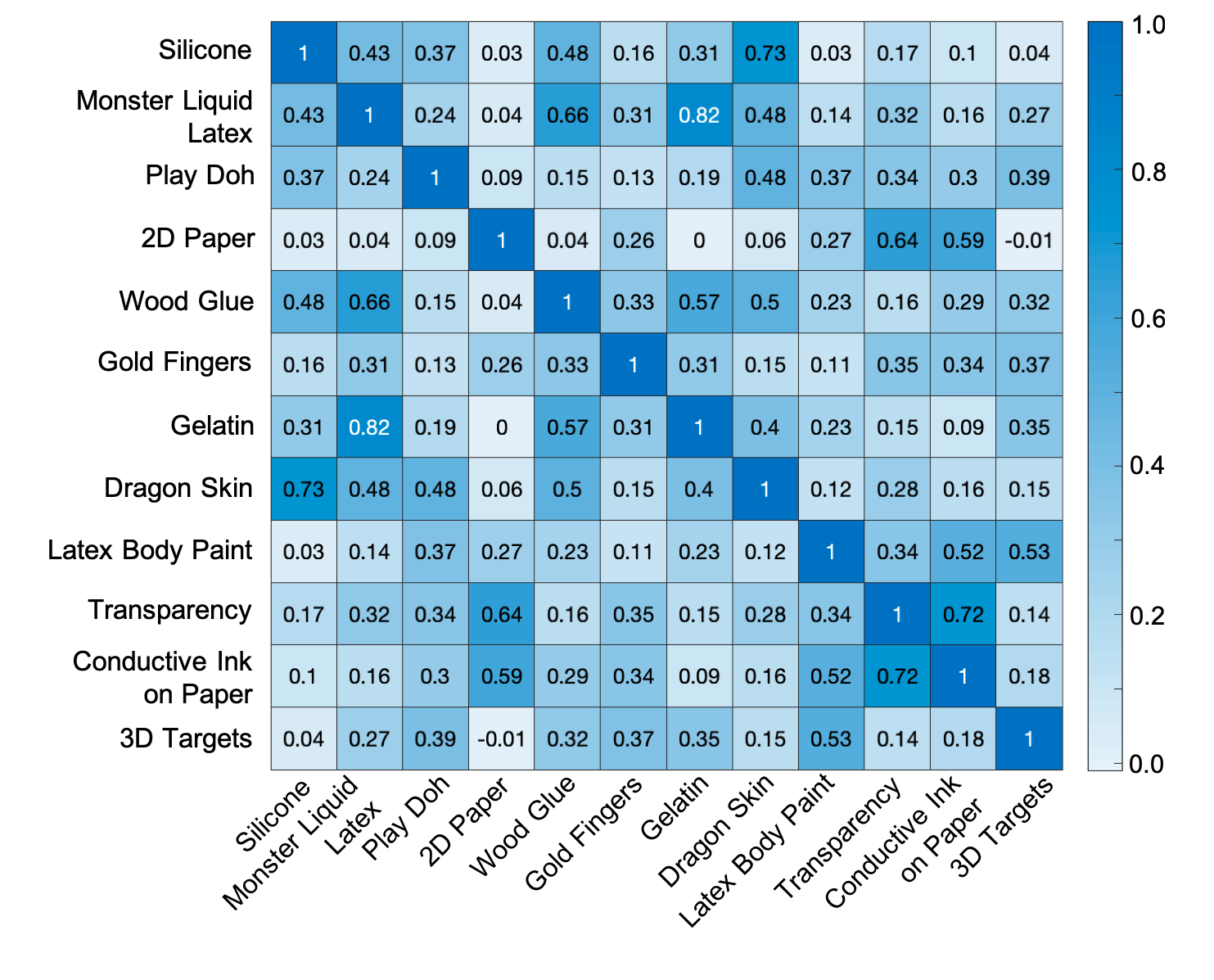}
\caption{Average Pearson correlation values between 12 PA materials based on the material characteristics.}
\label{fig:mat_corr}
\end{figure}

\begin{figure*}[htbp!]
\centering
\includegraphics[trim=0cm 0cm 0cm 0cm, width=\linewidth]{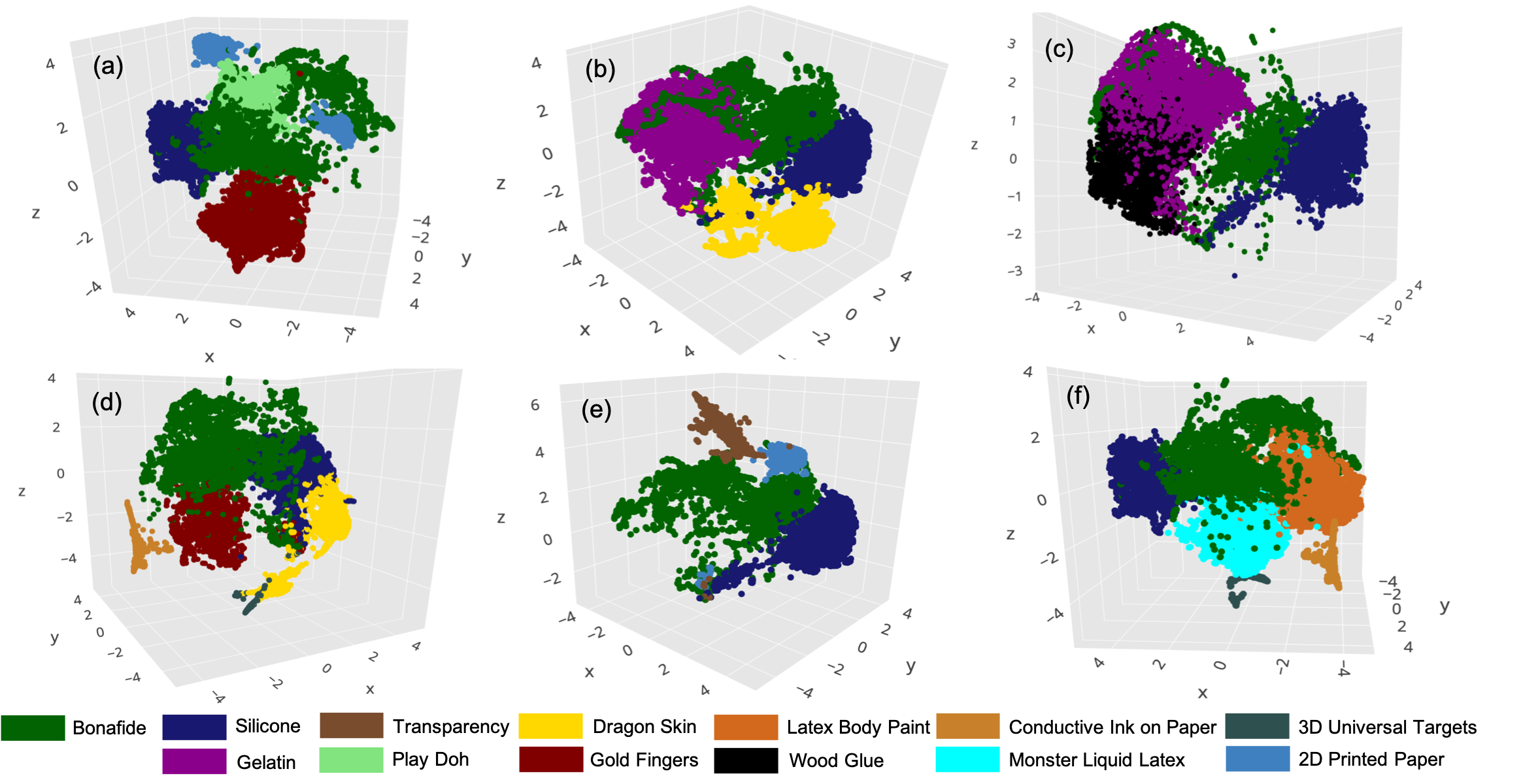}
\caption{Representation of bonafide and different subsets of PA materials in 3D t-SNE feature space from different angles selected to provide the best view. The bonafide (dark green) and silicone (navy blue) are included in all graphs for perspective.}
\label{fig:tsne}
\end{figure*}

%

\subsection{Feature Representation of Bonafide and PAs}
To explore the relationship between bonafide and different PA materials, we train a single multi-class MobileNet-v1 model to distinguish between $13$ classes, \textit{i.e.} bonafide and $12$ PA materials. The training split includes a set of randomly selected $100$ images or half the number of total images (whichever is lower) from each of bonafide and PA materials for a total of $1,102$ images. In a similar way, a test split is constructed from the remaining set of images for a total of $1,101$ images. This protocol is adopted to reduce the bias due to unbalanced nature of the training dataset. We extract the $1024$-dimensional feature vector from the bottleneck layer of the MobileNet-v1 network~\cite{howard2017mobilenets} and project it to 3 dimensions using t-SNE approach~\cite{maaten2008visualizing} (see Figure~\ref{fig:sampletsne}). Figures~\ref{fig:tsne} (a)-(f) present the representation of bonafide and different subsets of PA materials in 3D t-SNE feature space from different angles selected to provide a complete view. The Bonafide (dark green) and Silicone (navy blue) are included in all graphs for perspective. The 3D graph is generated using plotly library and is accessible at the link: \url{https://plot.ly/~icbsubmission/0/livepa-feature-space/#/}

\begin{figure}[t!]
\centering
\includegraphics[trim=1.5cm 0.5cm 1cm 0.8cm, width=0.85\linewidth]{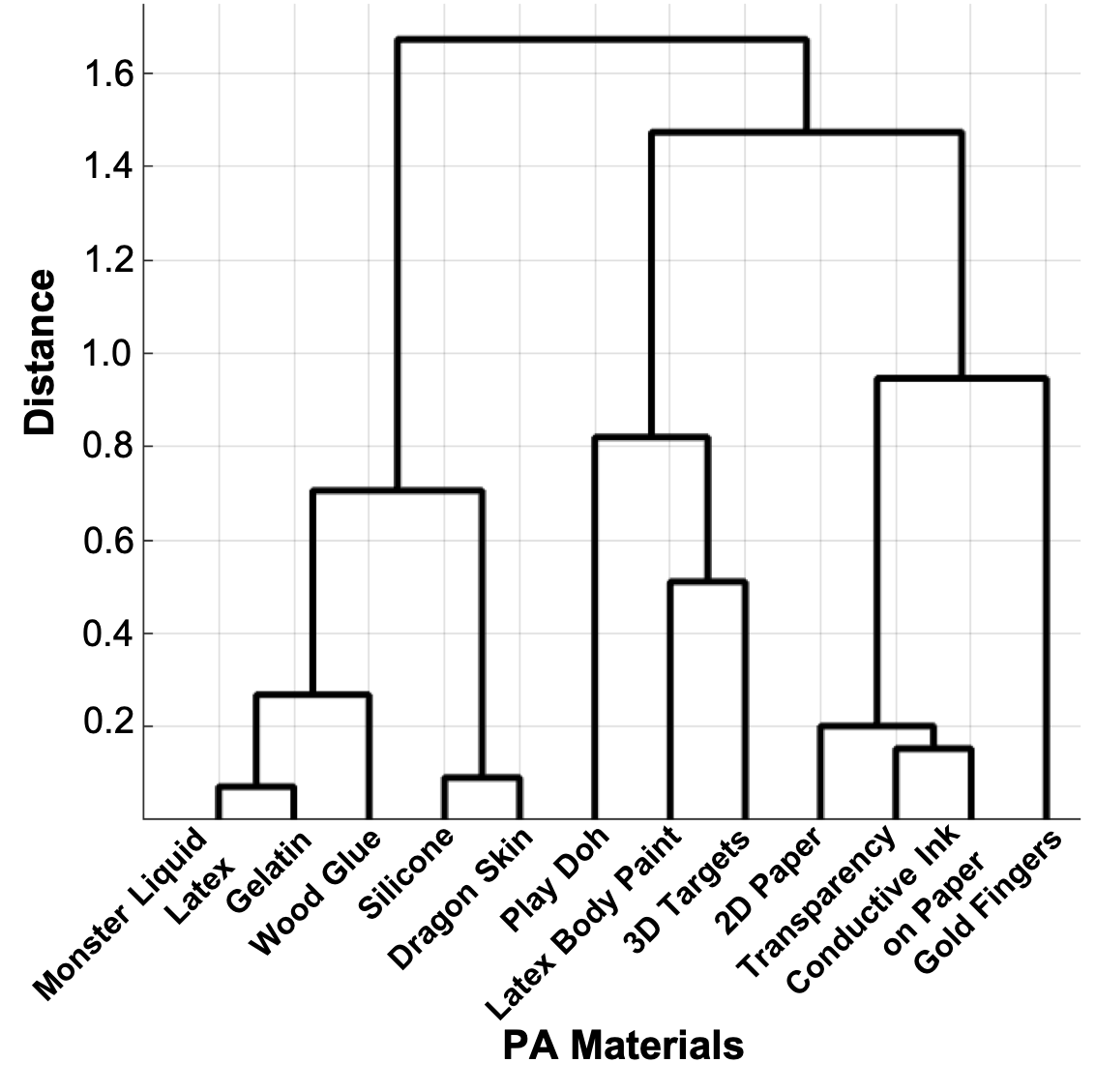}
\caption{A complete-link dendrogram representing the hierarchical (agglomerative) clustering of PAs based on the shared material characteristics.}
\label{fig:dendrogram}
\end{figure}

\begin{figure*}[t]
\centering
\includegraphics[trim=0cm 0cm 0cm 0cm, width=0.82\linewidth]{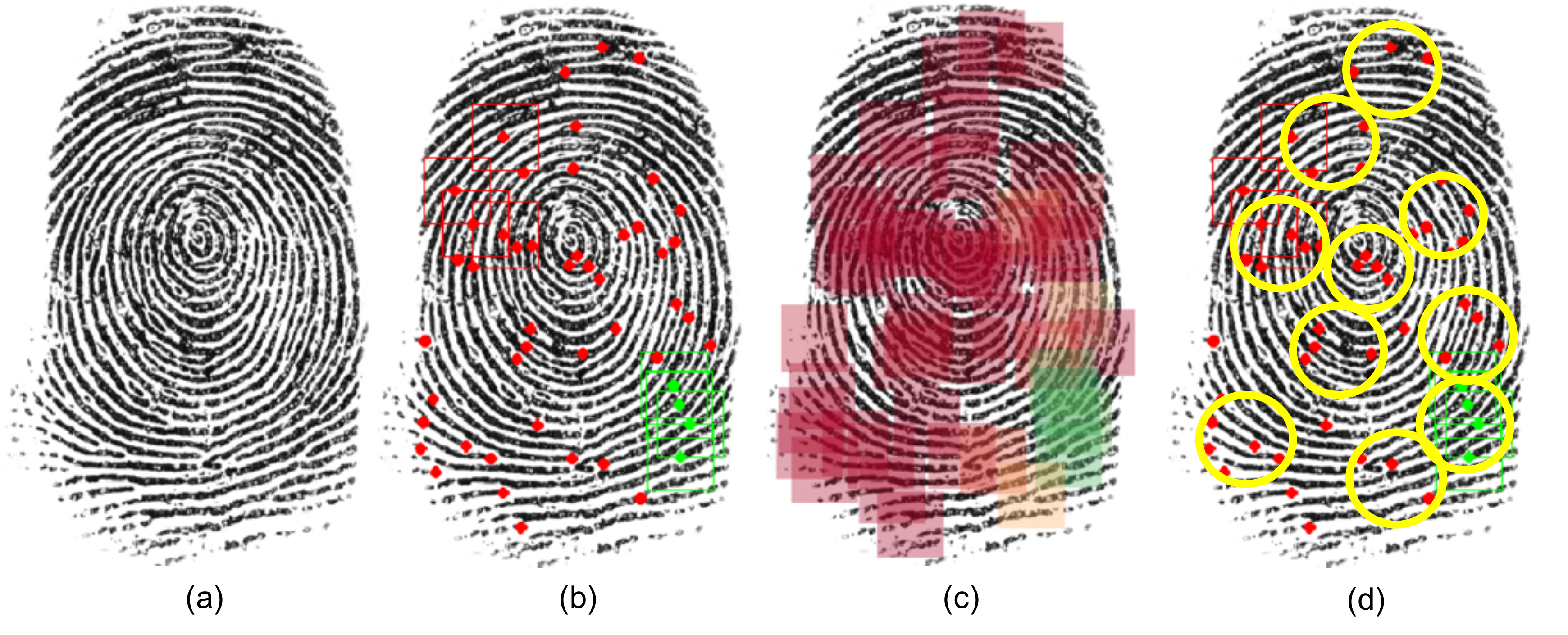}
\caption{Minutiae clustering. (a) fingerprint image; (b) extracted minutiae overlaid on (a); (c) $96 \times 96$ patches centered at each minutiae; (d) minutiae clustering using k-means (k is set to 10 here)}
\label{fig:minuClustering}
\vspace{-3.1mm}
\end{figure*}

\subsection{Representative set of PA Materials}

We utilize material characteristics and 3D t-SNE visualization to identify a set of representative materials to train a robust and generalized model. From the four material characteristics, two continuous (\textit{i.e.} optical characteristics) and two categorical (\textit{i.e.} mechanical characteristics), we compute four $12 \times 12$ correlation matrices. For the two continuous variables, we compute the Pearson correlation\footnote{MATLAB's \textit{corr} function is used to compute the Pearson correlation. \url{https://www.mathworks.com/help/stats/corr.html}} between all pairs of materials to generate two correlation matrices $C^{uvvis}$ and $C^{ftir}$. For the two categorical variables, if two PA materials $m_i$ and $m_j$ belong to the same category, we assign $C_{i,j} = 1$, else $C_{i,j} = 0$, to generate $C^{elastic}$ and $C^{moisture}$. The four correlation matrices corresponding to each of the four individual material characteristics, are averaged to generate the final correlation matrix $C^{material}$, such that $C^{material}_{i,j} = (C^{uvvis}_{i,j} + C^{ftir}_{i,j} + C^{elastic}_{i,j} + C^{moisture}_{i,j})/4$, (see Figure~\ref{fig:mat_corr}) which is utilized to perform complete-link hierarchical (agglomerative) clustering\footnote{We utilize MATLAB's \textit{linkage} and \textit{dendrogram} functions with parameters method=`complete' and metric=`correlation'.} of the 12 PA materials. Figure~\ref{fig:dendrogram} shows a complete-link dendrogram representing the hierarchical grouping of the 12 PA materials based on $C^{material}$. Based on the 3D t-SNE visualization and the hierarchical clustering of the 12 PA materials, we observe that:
\begin{itemize}
\item PA materials Silicone, Play Doh, Gelatin, and 2D Printed Paper are closest to Live fingerprints in the 3D t-SNE feature space compared to other materials. This explains why excluding any one of them resulted in poor generalization performance when tested against them. These PA materials appear in different clusters in the dendrogram (see Figures~\ref{fig:tsne} (a) and \ref{fig:dendrogram}).
\item PA material Dragon Skin is easily detected when Silicone is included in training as silicone lies between bonafide and Dragon Skin (see Figures~\ref{fig:tsne} (b) and (d)). These materials also lie in the same cluster indicating shared material characteristics.
\item PA material Transparency is easily distinguishable when 2D Printed Paper is included in training. In the t-SNE visualization, we observe that 2D Printed Paper forms two clusters, where one of the clusters is co-located with transparency (see Figures~\ref{fig:tsne} (a) and (e)).
\item PA materials Wood Glue and Gelatin are co-located in feature space assisting each other if included in training (see Figure~\ref{fig:tsne} (c)), but Gelatin is closer to Bonafide which explains its worse performance compared to Wood Glue. These materials also form a second level cluster in the dendrogram.
\item PA material Latex Body Paint lies between Bonafide and Conductive Ink on Paper, and PA material Monster Liquid Latex lies between Bonafide and 3D Universal Targets in 3D t-SNE visualization which could explain the high detection for Conductive Ink on Paper and 3D Universal Targets (see Figure~\ref{fig:tsne} (f)). However, these materials do not form a cluster until the last agglomeration step, indicating possibility of other material characteristics that could be further explored.
\end{itemize}

Based on these observations, we infer that a set of 6 PA materials, namely Silicone, 2D Paper, Play Doh, Gelatin, Latex Body Paint, and Monster Liquid Latex, almost covers the entire feature space around Bonafide (see Figure~\ref{fig:tsne}). A model trained using bonafide and these 6 PA materials achieved an average TDR = $89.76\%~\pm 6.97\%$ @ FDR = $0.2\%$ when tested on each of the remaining 6 materials. This performance is comparable to the average TDR = $90.97\% \pm 7.27\%$ @ FDR = $0.2\%$ when 11 PA materials are used for training, indicating no significant contribution provided by including the other 5 materials in training. We posit that the PAD performance against new materials can be estimated by analyzing its material characteristics instead of collecting large datasets for each of the new material. To address the second limitation of high resource requirement of CNN based PAD approach, we propose optimizations to Fingerprint Spoof Buster in the next section.


\begin{figure*}[t]
\centering
\includegraphics[trim=0cm 0cm 0cm 0cm, width=0.87\linewidth]{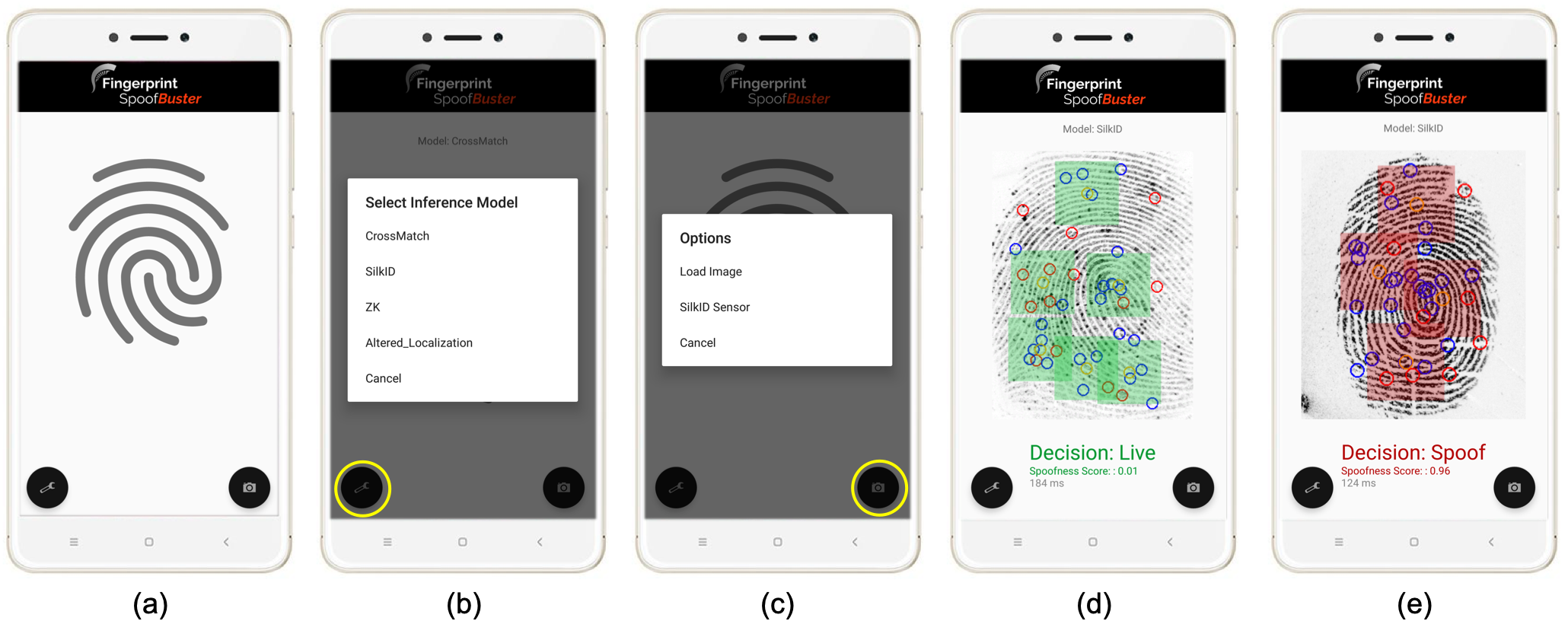}
\caption{User interface of the Android application, \textit{Fingerprint Spoof Buster Lite} shown in (a). It allows selection of inference model as shown in (b). User can load a fingerprint image from phone storage or capture a live scan from a fingerprint reader as shown in (c). The app executes PAD and displays the final decision along with highlighted local patches on the screen shown in (d) and (e).}
\label{fig:appscreens}
\vspace{-2.2mm}
\end{figure*}

\section{Fingerprint Spoof Buster Lite}


The Fingerprint Spoof Buster~\cite{chugh2018fingerprint} 
evaluates all local patches corresponding to the detected minutiae. The individual scores output by the CNN model for 
patch is averaged to produce a global spoofness score. 
The time required to evaluate a single patch utilizing MobileNet-v1 CNN model on a commodity smartphone, such as Redmi Note 4\footnote{https://www.gsmarena.com/xiaomi\_redmi\_note\_4-8531.php} (Qualcomm Snapdragon 625 64-bit Octa Core 2GHz Processor and 3GB RAM), is around $125$ms. This results in an average execution time of $4.3$ seconds per image (with an average no. of $35$ minutiae/image). Moreover, a MobileNet-v1 trained model in ProtoBuf (.pb) format takes around $13$MB. In order to reduce the memory and computation requirements for real-time operation on a commodity smartphone, we propose the following two optimizations:

\textit{Model Quantization:} \textit{Tensorflow-lite}\footnote{https://www.tensorflow.org/lite/} is used to convert the MobileNet-v1 (.pb) model to \textit{tflite} format, resulting in a light-weight and low-latency model with weights quantized to perform byte computations instead of float airthmetic. The resultant model size is only $3.2$MB and can execute PAD for a single patch on the same Redmi Note 4 smartphone in around $30$ms, approximately $80\%$ reduction in computation and memory requirements.

\textit{Reduce the number of patches:} It has been observed that minutiae points in a fingerprint image are distributed in a non-uniform manner~\cite{pankanti2002individuality}. This obviates the need for evaluating all minutiae-centered patches. We cluster the minutiae using K-means clustering (see Figure~\ref{fig:minuClustering} (d)), and then extract a single patch ($96 \times 96$) centered at the centroid of each cluster. A weight is assigned to the centroid based local patch corresponding to the number of minutiae belonging to that cluster. The final spoofness score is a weighted average of centroid-based local patches.

\begin{table}[t]
\centering
\caption{Detection time and PAD performance (TDR @ FDR = $0.2\%$) of Fingerprint Spoof Buster Lite.}
\label{tab:time}
\resizebox{\linewidth}{!}{
\begin{tabular}{ | >{\centering\arraybackslash}p{2cm} | >{\centering\arraybackslash}p{3.5cm} | >{\centering\arraybackslash}p{2.5cm} | } \hline
\textbf{\# Minutiae Clusters} & \textbf{Time Required (in ms) (Avg. $\pm$ s.d.) \hspace{2mm} } & \textbf{TDR (\%) @ FDR = 0.2\%} \\ \hline \hline
\textbf{$5$} & $148 \pm 19$ & $93.9 \pm 1.1$\\ \hline
\textbf{$\textbf{10}$} & $\textbf{302} \pm \textbf{15}$ & $\textbf{95.3} \pm \textbf{0.5}$\\ \hline
\textbf{$15$} & $447 \pm 13$ & $95.3 \pm 0.5$\\ \hline
\textbf{$20$} & $607 \pm 9$ & $95.3 \pm 0.6$\\ \hline
\textbf{$25$} &$736 \pm 21$ & $95.7 \pm 0.5$\\ \hline
\textbf{$30$} & $910 \pm 18$ & $95.7 \pm 0.4$\\ \hline
\textbf{All Minutiae (avg. = $\textbf{35}$)} & $\textbf{1554} \pm \textbf{11}$ & $\textbf{95.7} \pm \textbf{0.1}$\\ \hline
\end{tabular}
}
\vspace{-4mm}
\begin{flushleft}
{\footnotesize Note: RedMi Note 4 smartphone (Qualcomm Snapdragon 625 64-bit Octa Core 2GHz Processor and 3GB RAM) costs $\$150$.}
\end{flushleft}
\vspace{-5.2mm}
\end{table}


Apart from the above two optimizations, we modify the MobileNet-v1 network such that the input image size is $96 \times 96$, same as the patch size. Correspondingly, the kernel size used in the last average pool layer is reduced from $7 \times 7$ to $3 \times 3$. This reduces the time required to train the network on a dataset with around $100,000$ patches from $6$-$8$ hours to $2$-$2.5$ hours using a single NVIDIA GTX 1080Ti GPU, without any drop in PAD performance. We utilized the tensorflow-slim library\footnote{https://github.com/tensorflow/models/tree/master/research/slim} for our experiments.

Table~\ref{tab:time} presents the accuracy of Fingerprint Spoof Buster Lite (TDR (\%) @ FDR = 0.2\%) and the average time required to evaluate minutiae-based patches on Redmi Note 4. Since k-means clustering depends on the cluster initialization, we use 5-fold cross-validation and report average $\pm$ std for both the evaluation time and PAD performance. Table~\ref{tab:time} also shows that no more than 10 minutiae clusters are needed to maintain PAD performance, while reducing the computational requirement by almost $80\%$.

Given this reduction in resource requirements, an Android-based application (app) for Fingerprint Spoof Buster, called \textit{Fingerpint Spoof Buster Lite} was developed. The app provides an option to select an inference model trained on images from different fingerprint readers such as CrossMatch, SilkID\footnote{http://www.silkid.com/products/}, etc. as shown in Figure~\ref{fig:appscreens} (b). The app can evaluate a fingerprint image input by a fingerprint reader connected to the mobile phone via OTG cable. It also allows loading and evaluating an image from the phone storage/gallery (see Figure~\ref{fig:appscreens} (c)). The app displays the captured image, with extracted fingerprint minutiae overlaid on the fingerprint image. The golden circles represent the minutiae clusters. Local patches centered around the centroid of minutiae clusters are evaluated and highlighted based on the spoofness score. After evaluation, the app presents the final decision (Live / Spoof), spoofness score, and PA detection time (see Figures~\ref{fig:appscreens} (d) and (e)).

\section{Conclusions}
CNN based state-of-the-art PAD approaches suffer from two major limitations: (i) generalization against materials not seen during training, and (ii) high computation and memory requirements for execution. In this study, we have evaluated the generalization performance of a state-of-the-art PAD approach, namely Fingerprint Spoof Buster, using 12 different PA materials. We investigate a clustering of PA materials based on four material characteristics (two optical and two mechanical) and 3D t-SNE visualization, to explain the performance of cross-material experiments. We conclude that a subset of PA materials, namely Silicone, 2D Paper, Play Doh, Gelatin, Latex Body Paint, and Monster Liquid Latex are essential for training a robust PAD. Additionally, we proposed two optimizations to the state-of-the-art PAD and implemented an Android app solution that can run on a commodity smartphone (Xiaomi Redmi Note 4) without significant drop in performance and make a PA detection in real-time.

\section{Acknowledgement}
This research is based upon work supported in part by the Office of the Director of National Intelligence (ODNI), Intelligence Advanced Research Projects Activity (IARPA), via IARPA R\&D Contract No. 2017 - 17020200004. The views and conclusions contained herein are those of the authors and should not be interpreted as necessarily representing the official policies, either expressed or implied, of ODNI, IARPA, or the U.S. Government. The U.S. Government is authorized to reproduce and distribute reprints for governmental purposes notwithstanding any copyright annotation therein.

{\small
\bibliographystyle{ieee}
\bibliography{egbib}

\begin{thebibliography}{10}\itemsep=-1pt

\bibitem{antonelli2006fake}
A.~Antonelli, R.~Cappelli, D.~Maio, and D.~Maltoni.
\newblock Fake finger detection by skin distortion analysis.
\newblock {\em IEEE Transactions on Information Forensics and Security},
  1(3):360--373, 2006.

\bibitem{baldisserra2006fake}
D.~Baldisserra, A.~Franco, D.~Maio, and D.~Maltoni.
\newblock Fake fingerprint detection by odor analysis.
\newblock In {\em Proc. ICB}, pages 265--272. Springer, 2006.

\bibitem{biggio2012security}
B.~Biggio, Z.~Akhtar, G.~Fumera, G.~L. Marcialis, and F.~Roli.
\newblock Security evaluation of biometric authentication systems under real
  spoofing attacks.
\newblock {\em IET Biometrics}, 1(1):11--24, 2012.

\bibitem{cao2016hacking}
K.~Cao and A.~K. Jain.
\newblock {Hacking mobile phones using 2D Printed Fingerprints, MSU Tech.
  report, MSU-CSE-16-2}.
\newblock \url{https://www.youtube.com/watch?v=fZJI_BrMZXU}, 2016.

\bibitem{chugh2017fingerprint}
T.~Chugh, K.~Cao, and A.~K. Jain.
\newblock {Fingerprint Spoof Detection Using Minutiae-based Local Patches}.
\newblock In {\em Proc. IEEE IJCB}, pages 581--589, 2017.

\bibitem{chugh2018fingerprint}
T.~Chugh, K.~Cao, and A.~K. Jain.
\newblock {Fingerprint Spoof Buster: Use of Minutiae-centered Patches}.
\newblock {\em IEEE Transactions on Information Forensics and Security},
  13(9):2190--2202, 2018.

\bibitem{ding2016ensemble}
Y.~Ding and A.~Ross.
\newblock {An ensemble of one-class SVMs for fingerprint spoof detection across
  different fabrication materials}.
\newblock In {\em Proc. IEEE WIFS}, pages 1--6, 2016.

\bibitem{engelsma2018universal}
J.~J. Engelsma, S.~S. Arora, A.~K. Jain, and N.~G. Paulter.
\newblock {Universal 3D Wearable Fingerprint Targets: Advancing Fingerprint
  Reader Evaluations}.
\newblock {\em IEEE Transactions on Information Forensics and Security},
  13(6):1564--1578, 2018.

\bibitem{engelsma2018fingerprint}
J.~J. Engelsma, K.~Cao, and A.~K. Jain.
\newblock {Fingerprint Match in Box}.
\newblock In {\em IEEE BTAS}, 2018.

\bibitem{engelsma2018raspireader}
J.~J. Engelsma, K.~Cao, and A.~K. Jain.
\newblock Raspireader: Open source fingerprint reader.
\newblock {\em IEEE Transactions on Pattern Analysis \& Machine Intelligence},
  2018.

\bibitem{ghiani2013fingerprint}
L.~Ghiani, A.~Hadid, G.~L. Marcialis, and F.~Roli.
\newblock {Fingerprint liveness detection using Binarized Statistical Image
  Features}.
\newblock In {\em Proc. IEEE BTAS}, pages 1--6, 2013.

\bibitem{ghiani2017review}
L.~Ghiani, D.~A. Yambay, V.~Mura, G.~L. Marcialis, F.~Roli, and S.~A.
  Schuckers.
\newblock {Review of the Fingerprint Liveness Detection (LivDet) competition
  series: 2009 to 2015}.
\newblock {\em Image and Vision Computing}, 58:110--128, 2017.

\bibitem{gragnaniello2013fingerprint}
D.~Gragnaniello, G.~Poggi, C.~Sansone, and L.~Verdoliva.
\newblock {Fingerprint liveness detection based on Weber Local Image
  Descriptor}.
\newblock In {\em Proc. IEEE Workshop on Biometric Meas. Syst. Secur. Med.
  Appl.}, pages 46--50, 2013.

\bibitem{gragnaniello2015local}
D.~Gragnaniello, G.~Poggi, C.~Sansone, and L.~Verdoliva.
\newblock Local contrast phase descriptor for fingerprint liveness detection.
\newblock {\em Pattern Recognition}, 48(4):1050--1058, 2015.

\bibitem{hogan2015multiple}
J.~N. Hogan.
\newblock {Multiple reference OCT system}, 2015.
\newblock US Patent 9,113,782.

\bibitem{howard2017mobilenets}
A.~G. Howard, M.~Zhu, B.~Chen, D.~Kalenichenko, W.~Wang, T.~Weyand,
  M.~Andreetto, and H.~Adam.
\newblock Mobilenets: Efficient convolutional neural networks for mobile vision
  applications.
\newblock {\em arXiv preprint arXiv:1704.04861}, 2017.

\bibitem{isopad}
{International Standards Organization}.
\newblock {ISO/IEC 30107-1:2016, Information Technology---Biometric
  Presentation Attack Detection---Part 1: Framework}.
\newblock https://www.iso.org/standard/53227.html, 2016.

\bibitem{jang2017fingerprint}
H.-U. Jang, H.-Y. Choi, D.~Kim, J.~Son, and H.-K. Lee.
\newblock Fingerprint spoof detection using contrast enhancement and
  convolutional neural networks.
\newblock In {\em International Conference on Information Science and
  Applications}, pages 331--338. Springer, 2017.

\bibitem{lapsley1998anti}
P.~D. Lapsley, J.~A. Lee, D.~F. Pare~Jr, and N.~Hoffman.
\newblock Anti-fraud biometric scanner that accurately detects blood flow,
  1998.
\newblock US Patent 5,737,439, 1998.

\bibitem{maaten2008visualizing}
L.~V.~D. Maaten and G.~Hinton.
\newblock {Visualizing Data using t-SNE}.
\newblock {\em Journal of Machine Learning Research}, 9(Nov):2579--2605, 2008.

\bibitem{marasco2015survey}
E.~Marasco and A.~Ross.
\newblock A survey on antispoofing schemes for fingerprint recognition systems.
\newblock {\em ACM Computing Surveys}, 47(2):28, 2015.

\bibitem{marasco2012combining}
E.~Marasco and C.~Sansone.
\newblock Combining perspiration-and morphology-based static features for
  fingerprint liveness detection.
\newblock {\em Pattrn. Reco. Letters}, 33(9):1148--1156, 2012.

\bibitem{marcel2014handbook}
S.~Marcel, M.~S. Nixon, and S.~Z. Li.
\newblock {\em Handbook of Biometric Anti-Spoofing}.
\newblock Springer, 2014.

\bibitem{marcialis2010analysis}
G.~L. Marcialis, F.~Roli, and A.~Tidu.
\newblock Analysis of fingerprint pores for vitality detection.
\newblock In {\em Proc. 20th ICPR}, pages 1289--1292, 2010.

\bibitem{matsumoto2002impact}
T.~Matsumoto, H.~Matsumoto, K.~Yamada, and S.~Hoshino.
\newblock Impact of artificial gummy fingers on fingerprint systems.
\newblock In {\em Proc. SPIE}, volume 4677, pages 275--289, 2012.

\bibitem{menotti2015deep}
D.~Menotti, G.~Chiachia, A.~Pinto, W.~R. Schwartz, H.~Pedrini, A.~X. Falcao,
  and A.~Rocha.
\newblock Deep representations for iris, face, and fingerprint spoofing
  detection.
\newblock {\em IEEE Transactions on Information Forensics and Security},
  10(4):864--879, 2015.

\bibitem{mura2018livdet}
V.~Mura, G.~Orr{\`u}, R.~Casula, A.~Sibiriu, G.~Loi, P.~Tuveri, L.~Ghiani, and
  G.~L. Marcialis.
\newblock Livdet 2017 fingerprint liveness detection competition 2017.
\newblock In {\em Proc. ICB}, pages 297--302, 2018.

\bibitem{nogueira2016fingerprint}
R.~F. Nogueira, R.~de~Alencar~Lotufo, and R.~C. Machado.
\newblock {Fingerprint Liveness Detection Using Convolutional Neural Networks}.
\newblock {\em IEEE Transactions on Information Forensics and Security},
  11(6):1206--1213, 2016.

\bibitem{IARPAProject}
{ODNI, IARPA}.
\newblock {IARPA-BAA-16-04 (Thor)}.
\newblock https://www.iarpa.gov/index.php/research-programs/odin/odin-baa,
  2016.

\bibitem{pala2017deep}
F.~Pala and B.~Bhanu.
\newblock {Deep Triplet Embedding Representations for Liveness Detection}.
\newblock In {\em Deep Learning for Biometrics}, pages 287--307. Springer,
  2017.

\bibitem{pankanti2002individuality}
S.~Pankanti, S.~Prabhakar, and A.~K. Jain.
\newblock {On the Individuality of Fingerprints}.
\newblock {\em IEEE Transactions on Pattern Analysis \& Machine Intelligence},
  24(8):1010--1025, 2002.

\bibitem{perkampus2013uv}
H.-H. Perkampus.
\newblock {\em UV-VIS Spectroscopy and its Applications}.
\newblock Springer Science \& Business Media, 2013.

\bibitem{rattani2015open}
A.~Rattani, W.~J. Scheirer, and A.~Ross.
\newblock Open set fingerprint spoof detection across novel fabrication
  materials.
\newblock {\em IEEE Transactions on Information Forensics and Security},
  10(11):2447--2460, 2015.

\bibitem{rowe2006multispectral}
R.~K. Rowe and D.~P. Sidlauskas.
\newblock Multispectral biometric sensor, 2006.
\newblock US Patent 7,147,153.

\bibitem{smith2011fundamentals}
B.~C. Smith.
\newblock {\em {Fundamentals of Fourier Transform Infrared spectroscopy}}.
\newblock CRC press, 2011.

\bibitem{tolosana2018towards}
R.~Tolosana, M.~Gomez-Barrero, J.~Kolberg, A.~Morales, C.~Busch, and
  J.~Ortega-Garcia.
\newblock {Towards Fingerprint Presentation Attack Detection Based on
  Convolutional Neural Networks and Short Wave Infrared Imaging}.
\newblock In {\em Proc. BIOSIG}, pages 1--5, 2018.

\end{thebibliography}
}

\end{document}